\documentclass[letterpaper, 10 pt, conference]{ieeeconf}  % Comment this line out if you need a4paper

\IEEEoverridecommandlockouts                              % This command is only needed if 
\usepackage{graphics} % for pdf, bitmapped graphics files
\usepackage{graphicx}

\usepackage{subfig} % 专门用来排版子图的包
\usepackage{amsmath}

\usepackage{amssymb}

\usepackage{booktabs}

\usepackage{cite}

\usepackage{caption}

\usepackage{url}

\title{\LARGE \bf
Multi-Resolution Voxelized Map-Based Stereo Visual-Inertial Odometry
}

\author{Shuyi Pan$^{1}$, Hangtian Wang$^{1}$, Zhaoxing Zhang$^{1}$, Chengliang Zhang$^{1}$, Zikang Yuan$^{2*}$, Xin Yang$^{1}$%
\thanks{This research is supported by National Natural Science Foundation of China (U25B2045, 62472184) and the Fundamental Research Funds for the Central Universities.}
\thanks{$^{1}$Shuyi Pan, Hangtian Wang, Zhaoxing Zhang, Chengliang Zhang and Xin Yang are with Huazhong University of Science and Technology, Wuhan, 430074, China. (E-mail: {\tt\small shuyipan@hust.edu.cn; htwang@hust.edu.cn; zzx@hust.edu.cn; z\_chengliang@hust.edu.cn; xinyang2014@hust.edu.cn})}%
\thanks{$^{2}$Zikang Yuan$^{*}$ is with the AI Chip Center for Emerging Smart Systems (AC-CESS), InnoHK Centers, Hong Kong Science Park, Hong Kong SAR, China. (E-mail: {\tt\small zikangyuan@ust.hk})}%
\thanks{$^{*}$represents the corresponding author}
}

\begin{document}

\maketitle
\thispagestyle{empty}
\pagestyle{empty}

%%%%%%%%%%%%%%%%%%%%%%%%%%%%%%%%%%%%%%%%%%%%%%%%%%%%%%%%%%%%%%%%%%%%%%%%%%%%%%%%
\begin{abstract}

Incorporating prior maps significantly enhances the accuracy and robustness of pose estimation in visual-inertial odometry (VIO). However, the large data volume of such maps, combined with limited transmission bandwidth, makes it impractical to continuously load local maps onto an edge device. In this paper, we propose a multi-resolution prior map construction method and a corresponding map-based VIO system. The prior map is voxelized at multiple resolutions, with each voxel retaining only a single map point. During online VIO operation, a cone-shaped indexing strategy associates 2D features on the edge device with 3D map points. The cone’s intercept is determined by the distance from the current position to the 3D points, enabling the selection of the appropriate resolution level and the retrieval of the unique map point within the corresponding voxel via a 3D digital differential analyzer (DDA) algorithm. This approach minimizes both the volume of data required for transmission and the computational load during data association. Extensive experiments on two public datasets demonstrate that our system achieves accurate pose estimation while requiring minimal data transmission.

\end{abstract}

%%%%%%%%%%%%%%%%%%%%%%%%%%%%%%%%%%%%%%%%%%%%%%%%%%%%%%%%%%%%%%%%%%%%%%%%%%%%%%%%
\section{INTRODUCTION}

Visual-inertial odometry (VIO), which estimates 15-degree-of-freedom (15-DOF) states, finds wide applications including augmented reality (AR) and virtual reality (VR). Accurate state estimation in pure VIO is challenging, as it requires simultaneous localization and mapping. By incorporating a prior map, VIO can dedicate its full capacity to state estimation, leading to substantial improvements in precision and robustness. However, due to the limited computational resources of the edge device and limited transmission bandwidth, the data transmission scheme between the edge and the cloud, as well as the data association approach, remain problems worth exploring.

To address above challenges, some approaches \cite{bao2022robust,belkin2024localization,zuo2019visual,ye2020monocular,lu2019sharing,jeon2024ecar,yuan2024sr++,liu2026gemdepth,cheng2025monster,jia2020d,cheng2025monster++,wei2025decoupling,wei2025wavelet} transmit images and roughly estimated poses collected on the edge device to the cloud, where more refined global pose optimization is performed. The optimized global poses, along with local meshes of the surrounding area, are then sent back to the edge device. The edge device subsequently extracts sparse 3D points from the dense mesh to perform 2D-3D data association. Although this approach delegates time-consuming global pose optimization to the resource-rich cloud, it fails to address the transmission bandwidth constraint, since the surrounding area must still be transmitted to the edge device. Other approaches \cite{huang2020geometric,huang2020gmmloc} perform data association in the cloud and then transmit the associated 3D information to the local device in the form of Gaussian Mixture Model (GMM) components, thereby significantly reducing the demand on transmission bandwidth. However, this GMM-based approach is only suitable for small-scale environments and cannot meet the requirements of large-scale scenarios.

\begin{figure}[t]  % [thpb] 是位置参数：优先尝试 Top, Here, Page, Bottom
    \centering      % 让图片居中
    % width=3.0in 表示固定宽度,也可以用 width=0.9\linewidth (推荐,占行宽的90%)
    % figures/my_image.png 是你的图片路径
    \includegraphics[width=1.0\linewidth, trim=1.5cm 0.9cm 2.2cm 1.3cm, clip]{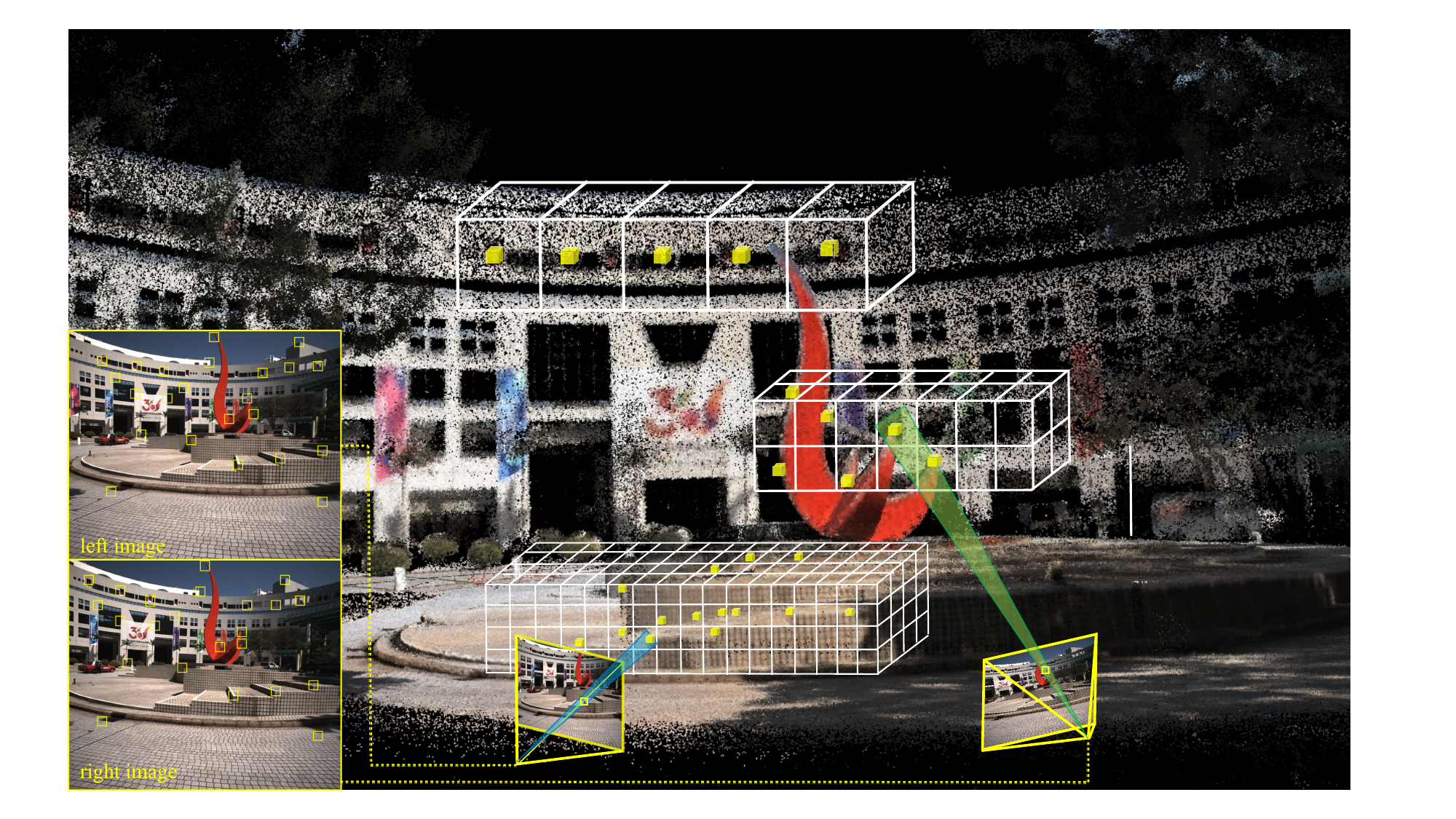}
     \caption{Illustration of multi-resolution voxelized prior map indexing for stereo VIO. White grids denote voxels at various resolutions. Depth-dependent 2D feature footprints determine the optimal voxel resolution, followed by a 3D-DDA ray traversal to retrieve associated map points, providing high-precision observation inputs for the state estimator.}
    \label{fig:indexing_diagram2} % 给图片起个“身份证号”,用于正文引用
    \vspace{-0.6cm} % 添加负间距,数值可以根据视觉效果微调（如 -0.2cm 到 -0.6cm）
\end{figure}

In this paper, we propose a multi-resolution prior map construction method and a corresponding map-based VIO framework. The prior map is voxelized at multiple resolutions, with each voxel retaining only a single map point. The system selects an appropriate voxel resolution of the prior map based on the physical size that each pixel represents in the real world. After dynamically adjusting the indexing range, the system employs a 3D digital differential analyzer (3D-DDA) algorithm \cite{amanatides1987fast,zhang2025leveraging} to perform ray traversal and retrieve the corresponding 3D points from the voxelized prior map, thereby associate 2D features on the edge device with prior 3D map points (as illustrated in Fig. \ref{fig:indexing_diagram2}). This approach not only minimizes the amount of data that needs to be transmitted between the cloud and the edge, but also reduces the computational load required for data correlation. At the edge VIO, we employ a two-stage state estimation approach \cite{geneva2020openvins, yuan2025voxel,cheng2024coatrsnet,cheng2022region} to compute the 15-DOF state. In the first stage, a coarse state is solved using 3D points retrieved from the multi-resolution voxelized prior map. In the second stage, the state is further refined by incorporating associated image features and prior 3D map points, yielding more accurate results. This two-stage state estimation strategy can significantly prevent the independent variables from falling into local minima. Analytical evaluation shows that our method requires extremely low data transmission between the edge and the cloud. Furthermore, experimental results on two public datasets (including indoor and outdoor scenarios) demonstrate that the proposed approach outperforms the state-of-the-art frameworks in terms of pose estimation accuracy. 

In summary, our contributions can be summarized as follows: 1) We introduce multi-resolution voxelized prior LiDAR maps into the VIO framework and develop a dynamic ray-based indexing strategy using a 3D-DDA, enabling efficient and accurate retrieval of structural information from the prior map. 2) We develop a map-based VIO framework based on the proposed multi-resolution voxelized prior LiDAR maps, which minimizes both the volume of data required for transmission and the computational load during data association. 3) We have released the source code of this work for the development of the community\footnote{\url{https://github.com/PANshuyi/MR-Voxel-SVIO}}.

The rest of this paper is structured as outlined below. A review of related work is presented in Sec. \ref{Related Work}. Sec. \ref{MULTI-RESOLUTION VOXEL INDEXING} describes the proposed multi-resolution voxel indexing and data association strategy. The methodology is detailed in Sec. \ref{Methodology}. Sec. \ref{EXPERIMENTS} presents the experimental evaluation, and Sec. \ref{CONCLUSIONS} concludes the paper.

\section{RELATED WORK}
\label{Related Work}
Following \cite{bao2022robust}, depending on whether global localization poses or the structural information of pre-built maps is utilized as additional measurements to improve the real-time accuracy of VIO, we can categorize the prior-map-based approaches into loosely-coupled and
tightly-coupled ones.

\textit{Loosely-Coupled Approaches:} In loosely coupled approaches, VIO/VO operates independently and relies on the prior map solely for low-frequency global pose corrections. Since map structural information is not directly incorporated into the state update, these methods intermittently mitigate accumulated drift rather than improving real-time estimation accuracy. Qin et al. \cite{qin2018relocalization} match local features against a sparse pose graph map to establish loop closures, triggering a global 4-DoF pose graph optimization to correct accumulated drift. Similar to \cite{qin2018relocalization}, Yamaguchi et al. \cite{yamaguchi2020global} estimate the scale and transformation between a VO graph and a global SfM map via 7-DoF optimization. Platinsky et al. \cite{platinsky2020collaborative} leverage image-based localization to refine the local-to-global transformation within a sliding window.

\textit{Tightly-Coupled Approaches:} Tightly coupled approaches incorporate global localization poses or map structural information into the state update as additional measurements to improve the real-time accuracy of VIO. Zuo et al. \cite{zuo2019visual} integrate NDT-based registration between stereo reconstructions and a prior LiDAR map into an MSCKF framework, which struggles to model localization uncertainty at low update frequencies. Caselitz et al. \cite{caselitz2016monocular} recover a 7-DoF transformation by aligning monocular images with a 3D LiDAR map via ICP and local point distribution analysis. Lü et al. \cite{lu2019sharing} fuse visual SLAM and prior LiDAR maps via vertical plane anchors, enforcing coplanarity constraints in global bundle adjustment to reduce estimation errors. Lynen et al. \cite{lynen2015get} develop a system coupled with an SfM-based 3D point cloud. Although using compact feature representations, it has limited scalability to large scenes and is sensitive to environmental changes. Ding et al. \cite{ding2018laser} present a LiDAR-assisted visual–inertial localization framework that aligns a local visual map with a prior LiDAR map via sliding-window optimization, while Kim et al. \cite{kim2018stereo} generate a local point cloud from the current disparity map to directly track the prior LiDAR map. While Huang et al. \cite{huang2020geometric,huang2020gmmloc,cheng2024adaptive} transmit cloud-associated 3D information as GMM components to reduce bandwidth, this representation is inherently restricted to small-scale environments and scales poorly to large scenarios. Bao et al. \cite{bao2022robust} offload global pose optimization to the cloud by transmitting images and roughly estimated poses. The cloud then returns optimized global poses and local dense meshes, from which the edge device extracts sparse 3D points to perform 2D-3D data association. Ye et al. \cite{ye2020monocular} improve robustness to pose errors by integrating rendered surfel constraints into the direct photometric error, this approach underutilizes the prior map and remains sensitive to map noise. Belkin et al. \cite{belkin2024localization} converts stereo dense depth into pseudo point clouds and directly registers them with a LiDAR map via ICP, avoiding information loss from feature extraction. Jeon et al. \cite{jeon2024ecar} offloads complex loop closure and back-end optimization tasks to the cloud and downloads lightweight local graph structure on demand, thereby significantly reducing network traffic.

Inspired by these works, to overcome the limitations of existing methods regarding computational resources on edge devices and transmission bandwidth between the cloud and the edge, we propose a multi-resolution prior map construction method and a corresponding map-based VIO framework. By dynamically matching the voxel resolution to the real-world physical size represented by 2D features and utilizing 3D-DDA ray traversal for point retrieval, our system achieves precise 2D-to-3D data association while minimizing both the required transmission bandwidth and the computational load on the edge device.
% \begin{figure*}[t]  % [thpb] 是位置参数：优先尝试 Top, Here, Page, Bottom
%     \centering      % 让图片居中
%     % width=3.0in 表示固定宽度,也可以用 width=0.9\linewidth (推荐,占行宽的90%)
%     % figures/my_image.png 是你的图片路径
%     \includegraphics[width=0.9\linewidth, trim=1cm 0cm 0.2cm 0cm, clip]{figure/fig3.pdf}
%     \caption{Overview of our MR-Voxel-SVIO which consists of four modules: an offline processing engine, an image processing module, a MSCKF based odometry module and a point cloud indexing and transmission module.}
%     \label{fig:indexing_diagram4} % 给图片起个“身份证号”,用于正文引用
%     \vspace{-0.4cm} % 添加负间距,数值可以根据视觉效果微调（如 -0.2cm 到 -0.6cm）
% \end{figure*}

\begin{figure*}[t]  % [thpb] 是位置参数：优先尝试 Top, Here, Page, Bottom
    \centering      % 让图片居中
    % width=3.0in 表示固定宽度,也可以用 width=0.9\linewidth (推荐,占行宽的90%)
    % figures/my_image.png 是你的图片路径
    \includegraphics[width=0.9\linewidth, trim=0.2cm 0cm 0.1cm 0.2cm, clip]{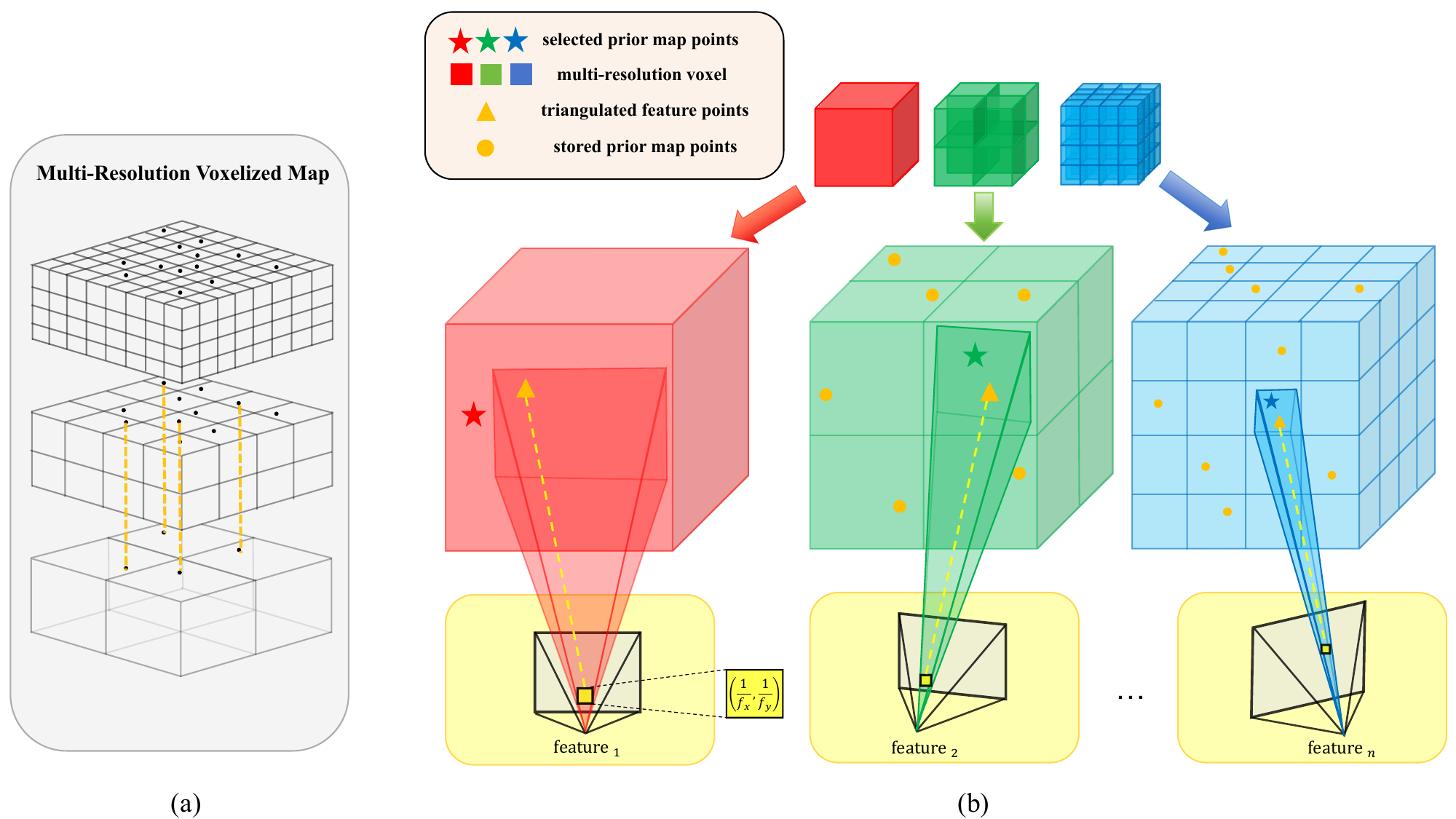}
    \caption{Illustration of the proposed multi-resolution voxel indexing and data association. (a) Preconstructed multi-resolution voxelized prior maps. The yellow dashed lines indicate corresponding points across different resolutions. (b) Online map querying based on the cone intercept for resolution selection, followed by 3D-DDA ray traversal to establish 2D–3D correspondences.}

    \label{fig:indexing_diagram3} % 给图片起个“身份证号”,用于正文引用
    \vspace{-0.4cm} % 添加负间距,数值可以根据视觉效果微调（如 -0.2cm 到 -0.6cm）
\end{figure*}

% \begin{figure*}[t]  % [thpb] 是位置参数：优先尝试 Top, Here, Page, Bottom
%     \centering      % 让图片居中
%     % width=3.0in 表示固定宽度,也可以用 width=0.9\linewidth (推荐,占行宽的90%)
%     % figures/my_image.png 是你的图片路径
%     \includegraphics[width=0.9\linewidth, trim=0.2cm 0cm 0.1cm 0.2cm, clip]{figure/fig2.pdf}
%     \caption{Illustration of the proposed multi-resolution voxel indexing and data association. (a) Preconstructed multi-resolution voxelized prior maps. The yellow dashed lines indicate corresponding points across different resolutions. (b) Online map querying based on the cone intercept for resolution selection, followed by 3D-DDA ray traversal to establish 2D–3D correspondences.}

%     \label{fig:indexing_diagram3} % 给图片起个“身份证号”,用于正文引用
%     \vspace{-0.4cm} % 添加负间距,数值可以根据视觉效果微调（如 -0.2cm 到 -0.6cm）
% \end{figure*}

\section{Multi-Resolution Voxel Indexing and Data Association}
\label{MULTI-RESOLUTION VOXEL INDEXING}
Fig. \ref{fig:indexing_diagram3} illustrates the implementation of the proposed multi-resolution voxelized map selection strategy. As shown in Fig. \ref{fig:indexing_diagram3}(a), the prior LiDAR map is constructed offline at multiple voxel resolutions. The map space is partitioned into voxels of different sizes (0.01, 0.02, and 0.03 m for indoor scenarios; 0.05, 0.10, 0.15, and 0.20 m for outdoor scenarios), and each voxel retains only the point with the maximum intensity gradient to ensure structural distinctiveness while maintaining compactness, thereby minimizing both the volume of data required for transmission and the computational load during the data association process. Furthermore, this highest-gradient point is consistently aligned and preserved across all resolution levels. The rationale behind this configuration is twofold: 1) to align voxel dimensions with the physical spatial scales of image pixels across typical depth ranges, and 2) to ensure robust system performance across diverse scenarios, as validated by extensive experiments. 

As illustrated in Fig. \ref{fig:indexing_diagram3}(b), during the online phase, the depths of stereo feature points are first estimated via triangulation. By incorporating camera intrinsics ($f_x, f_y$) to establish the pixel dimensions, the feature footprint is modeled as a projection pyramid, whose intercept at the estimated depth dictates the corresponding real-world spatial scale through geometric similarity. The voxel resolution level that best accommodates this scale is then dynamically selected. To avoid the redundant sampling and missed intersections inherent to fixed-step ray casting across multi-resolution voxels, the 3D-DDA algorithm \cite{amanatides1987fast} is employed. Starting within the vicinity of the estimated depth, it dynamically adjusts the indexing range and incrementally traverses the line of sight until intersecting a valid prior map voxel or exceeding the predefined search range, naturally adapting to the selected voxel resolution. Upon locating a valid voxel, the unique prior point residing within is extracted and, subject to a rigorous error validation, robustly associated with the corresponding 2D feature, ensuring high accuracy of 2D-to-3D correspondences and minimizing the computational load during data association.

% \begin{figure*}[t]  % [thpb] 是位置参数：优先尝试 Top, Here, Page, Bottom
%     \centering      % 让图片居中
%     % width=3.0in 表示固定宽度,也可以用 width=0.9\linewidth (推荐,占行宽的90%)
%     % figures/my_image.png 是你的图片路径
%     \includegraphics[width=0.9\linewidth, trim=1cm 0cm 0.2cm 0cm, clip]{figure/fig3.pdf}
%     \caption{Overview of our MR-Voxel-SVIO which consists of four modules: an offline processing engine, an image processing module, a MSCKF based odometry module and a point cloud indexing and transmission module.}
%     \label{fig:indexing_diagram4} % 给图片起个“身份证号”,用于正文引用
%     \vspace{-0.4cm} % 添加负间距,数值可以根据视觉效果微调（如 -0.2cm 到 -0.6cm）
% \end{figure*}
% \section{Methodology}
% \label{Methodology}

\begin{figure*}[t]  % [thpb] 是位置参数：优先尝试 Top, Here, Page, Bottom
    \centering      % 让图片居中
    % width=3.0in 表示固定宽度,也可以用 width=0.9\linewidth (推荐,占行宽的90%)
    % figures/my_image.png 是你的图片路径
    \includegraphics[width=0.9\linewidth, trim=1cm 0cm 0.2cm 0cm, clip]{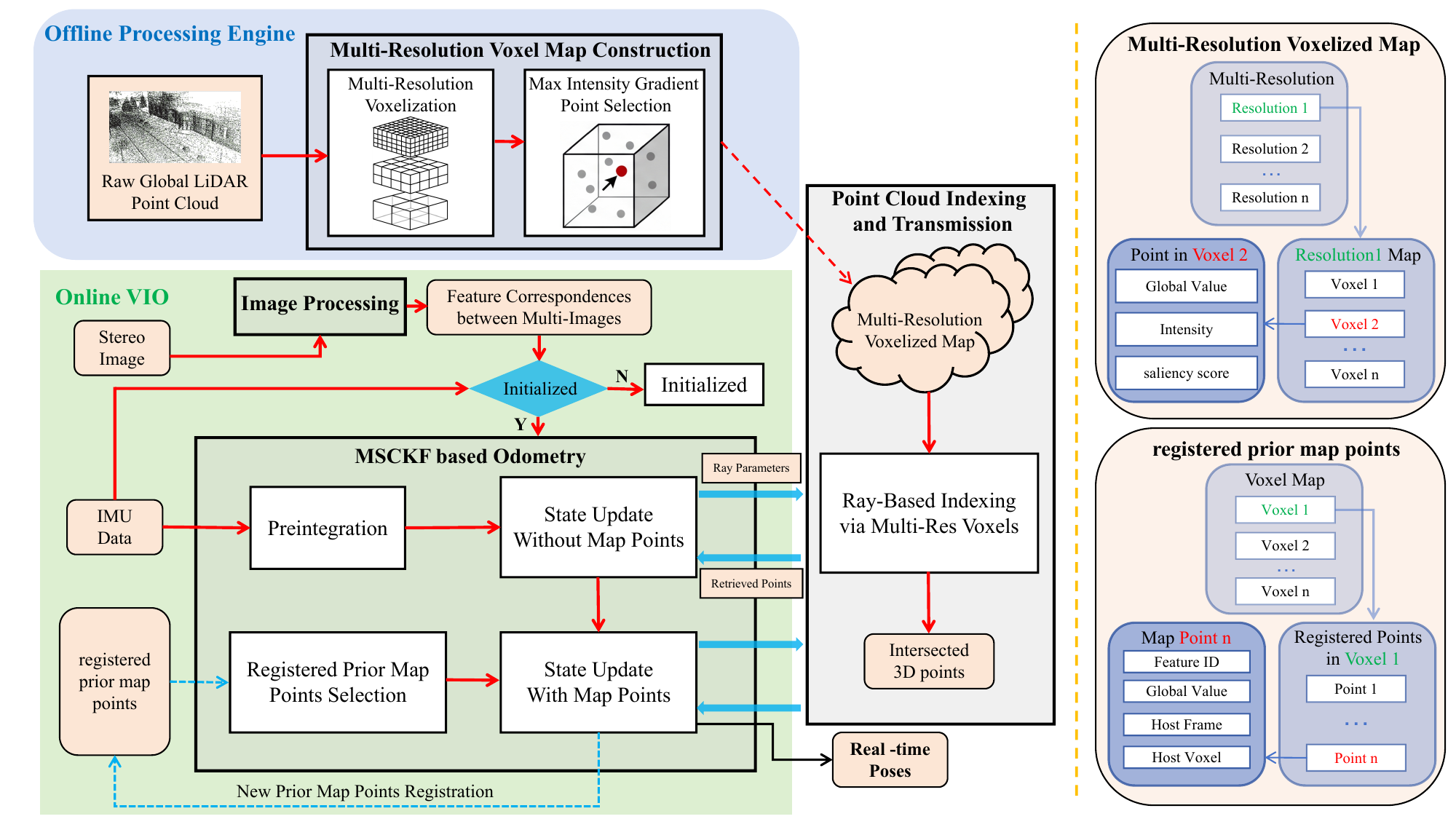}
    \caption{Overview of our MR-Voxel-SVIO which consists of four modules: an offline processing engine, an image processing module, a MSCKF based odometry module, a point cloud indexing and transmission module.}
    \label{fig:indexing_diagram4} % 给图片起个“身份证号”,用于正文引用
    \vspace{-0.4cm} % 添加负间距,数值可以根据视觉效果微调（如 -0.2cm 到 -0.6cm）
\end{figure*}

\section{OUR SYSTEM MR-VOXEL-SVIO}
\label{Methodology}
\subsection{system overview} 
Fig. \ref{fig:indexing_diagram4} illustrates the overall framework of the proposed system. Upon receiving stereo images, the image processing module efficiently conducts continuous feature tracking and stereo matching to establish robust multi-view feature correspondences. After successful initialization, the system enters the MSCKF-based odometry module, where state prediction is carried out via IMU preintegration. Then, a ray-based indexing of 2D feature correspondences across all frames in the sliding window is performed to retrieve global 3D positions, followed by an update of the state for these frames. Subsequently, visual constraints are formulated utilizing suitable map points selected from the registered prior map points and their corresponding 2D features across the sliding window, leading to a further refinement of the states for all frames and the global 3D value for selected suitable map points. Upon completion of the state update, the system indexes the 3D positions for new features in the current frame and establishes 2D-3D associations, incorporating them into the registered prior map points.

\subsection{Offline Multi-Resolution Voxel Map Construction}
To provide reliable global constraints for the online visual-inertial odometry, a compact prior map is constructed offline from a raw global LiDAR point cloud. The prior map is represented as a multi-resolution voxel structure $\mathcal{M} = \{\mathcal{M}_{r_1}, \mathcal{M}_{r_2}, \dots, \mathcal{M}_{r_n}\}$, where each layer $\mathcal{M}_{r_i}$ consists of voxels with a specific resolution $r_i$. To facilitate efficient data association and minimize data transmission latency, the continuous space is discretized such that each individual voxel $\mathcal{V}$ retains only a single representative LiDAR point $\mathbf{p}^* = [\boldsymbol{x}, I, S]^\top$, where $\mathbf{x} \in \mathbb{R}^3$ denotes the 3D position, $I \in \mathbb{R}$ represents the intensity value, and $S \in \mathbb{R}$ signifies the photometric saliency score $S_{photo}$. To maintain consistency with the preference of KLT tracking for features with significant intensity gradients, LiDAR points with high photometric saliency are prioritized during the selection process. This alignment ensures that the prior map points correspond to the most distinctive visual textures, thereby enhancing the reliability of cross-modal data association.

For each candidate point $\mathbf{p}_i$ within a voxel, its photometric saliency $S_{photo}(\mathbf{p}_i)$ is evaluated by computing a normal-weighted intensity gradient within its $K$-nearest neighbors $\mathcal{N}_i$:
% \begin{equation}
\begin{gather}
    S_{photo}(\mathbf{p}_i) \!=\! \max_{\mathbf{p}_j \in \mathcal{N}_i} \! \bigg[ \frac{|I_i \!-\! I_j|}{\|\mathbf{p}_i \!-\! \mathbf{p}_j\|} \Big( 1 \!+\! \alpha \Big| \mathbf{n}_i \!\cdot\! \frac{\mathbf{p}_j \!-\! \mathbf{p}_i}{\|\mathbf{p}_j \!-\! \mathbf{p}_i\|} \Big| \Big) \bigg]
\end{gather}
% \end{equation}
where $\mathbf{n}_i$ is the surface normal estimated from the local neighborhood. The scaling parameter $\alpha$ is utilized to prioritize 3D structural edges by amplifying intensity variations associated with depth discontinuities.

Ultimately, the optimal representative point $\mathbf{p}^*$ for each voxel is uniquely determined by maximizing this photometric saliency score:
\begin{gather}
    \mathbf{p}^* = \arg\max_{\mathbf{p}_i \in \mathcal{V}} S_{photo}(\mathbf{p}_i)
\end{gather}

Consequently, this strategy yields a sparse yet visually representative prior map, establishing a robust foundation for efficient online ray-based indexing, reliable cross-modal data association, and high-precision state refinement.

\subsection{Image Processing}
Upon receiving new stereo image, existing features are temporally tracked in the left image via KLT optical flow \cite{lucas1981iterative}, with untracked points promptly discarded. To compensate for inevitable feature loss and maintain a uniform spatial distribution, successfully tracked points are temporarily masked out. New features are then efficiently extracted from the unmasked regions using an improved DSO-based multi-resolution grid strategy \cite{engel2017direct}, which prioritizes pixels with the highest local intensity gradients. Finally, robust stereo correspondences are established by applying KLT optical flow across the current left and right images.

\subsection{Initialization}
Following \cite{chen2021rnin}, a motion-aware initialization strategy is employed to establish the initial system state. In static scenarios, initialization is achieved via gravity alignment and IMU bias estimation. Conversely, in dynamic environments, IMU preintegration is utilized to recover the metric scale, initial velocity, and gravity vector, which are subsequently refined through Ceres-based nonlinear optimization. Upon satisfying the convergence criteria with sufficient feature observations, the system seamlessly transitions to the standard state estimation pipeline. Furthermore, as relocalization falls outside the scope of this work, a one-time global coordinate alignment is executed immediately post-initialization. The local odometry frame is aligned with the global coordinate frame utilizing the initial ground-truth pose. It is strictly emphasized that no subsequent ground-truth information is accessed during the online operation.

\subsection{MSCKF Based Odometry}
The system state vector is defined as
\begin{gather}
\boldsymbol{x}_k =
\left[
{\boldsymbol{x}_I}^{\top}\;
{\boldsymbol{x}_C}^{\top}\;
{\boldsymbol{x}_M}^{\top}\;
{\boldsymbol{x}_{Y_0}}^{\top}\;
{\boldsymbol{x}_{Y_1}}^{\top}\;
t_I^C
\right]^{\top}
\end{gather}
where $\boldsymbol{x}_I$ denotes the current inertial navigation state, 
$\boldsymbol{x}_C$ contains $n$ pose clones within the sliding window, 
$\boldsymbol{x}_M$ represents the global 3D coordinates of $m$ selected map points, 
$\boldsymbol{x}_{Y_0}$ and $\boldsymbol{x}_{Y_1}$ are the stereo extrinsic and intrinsic parameters, 
and $t_I^C$ is the temporal offset between the IMU and camera clocks.
% \vspace{-0.1cm} % 添加负间距,数值可以根据视觉效果微调（如 -0.2cm 到 -0.6cm）
\begin{gather}
\boldsymbol{x}_I =
\left[
{\mathbf{q}_G^{I_k}}^{\top}\;
{\mathbf{t}_{I_k}^{G}}^{\top}\;
{\mathbf{v}_{I_k}^{G}}^{\top}\;
{\mathbf{b}_{{\omega}_k}}^{\top}\;
{\mathbf{b}_{a_k}}^{\top}
\right]^{\top}\\
\boldsymbol{x}_C =
\left[
{\mathbf{q}_{I_{k-1}}^{G}}^{\top}\;
{\mathbf{t}_{I_{k-1}}^{G}}^{\top}\;
\ldots\;
{\mathbf{q}_{I_{k-n}}^{G}}^{\top}\;
{\mathbf{t}_{I_{k-n}}^{G}}^{\top}\;
\right]^{\top}\\
\boldsymbol{x}_M =
\left[
{\mathbf{p}_1}^{\top}\;
{\mathbf{p}_2}^{\top}\;
\ldots\;
{\mathbf{p}_m}^{\top}
\right]^{\top}\\
\boldsymbol{x}_{Y_0} =
\left[
{\mathbf{q}_{C_0}}^{\top}\;
{\mathbf{t}_I^{C_0}}^{\top}\;
{\mathbf{F}_0}^{\top}\;
{\mathbf{D}_0}^{\top}
\right]^{\top}\\
\boldsymbol{x}_{Y_1} =
\left[
{\mathbf{q}_{C_1}}^{\top}\;
{\mathbf{t}_I^{C_1}}^{\top}\;
{\mathbf{F}_1}^{\top}\;
{\mathbf{D}_1}^{\top}
\right]^{\top}
\end{gather}
Here, $(\cdot)^C$, $(\cdot)^I$, and $(\cdot)^G$ denote quantities expressed in the camera, IMU, and global frames, respectively. The global frame is defined to be aligned with that of the prior map. The quaternion $\mathbf{q}$ represents the rotation, which can be converted to the corresponding rotation matrix $\mathbf{R}$. The vectors $\mathbf{t}$ and $\mathbf{v}$ denote the translation and linear velocity respectively. The bias terms $\mathbf{b}_{\omega}$ and $\mathbf{b}_a$ correspond to the gyroscope and accelerometer biases. The 3D position of registered prior map point in the global frame is denoted by $\mathbf{p}$. The intrinsic parameters of each camera are represented by the focal length vector $\mathbf{F}$ and the distortion parameter vector $\mathbf{D}$. We adopt the following generalized addition and subtraction operators for state updates. In addition, following \cite{geneva2020openvins, yuan2025voxel}, we employ a two-stage state estimation approach to compute the 15-DOF state.
\begin{gather}
    \mathbf{R} \boxplus \boldsymbol{\theta} = \mathbf{R} \mathrm{Exp}(\boldsymbol{\theta}); \mathbf{R}_1 \boxminus \mathbf{R}_2 = \log \left( \mathbf{R}_2^{\mathrm{\top}} \mathbf{R}_1 \right) \label{eq:manifold_ops} \\
    \mathbf{a} \boxplus \mathbf{b} = \mathbf{a} + \mathbf{b}; \mathbf{a} \boxminus \mathbf{b} = \mathbf{a} - \mathbf{b} \nonumber
\end{gather}
where $\mathbf{R}, \mathbf{R}_1, \mathbf{R}_2 \in SO(3)$ denote the rotation matrices in the Special Orthogonal group, $\boldsymbol{\theta} \in so(3)$ is the rotation vector in the corresponding Lie algebra, and $\mathbf{a}, \mathbf{b} \in \mathbb{R}^n$ are standard vectors in Euclidean space.

% \textit{1) IMU Propagation:} Upon receiving IMU measurements (linear acceleration $\boldsymbol{a}_k$ and angular velocity $\boldsymbol{\omega}_k$), the nominal inertial state is propagated from timestep $k-1$ to $k$:
% \begin{equation} \label{eq:state_predict}
% \hat{\boldsymbol{x}}_{k|k-1}^I = f \left( \hat{\boldsymbol{x}}_{k-1|k-1}^I, \boldsymbol{a}_k, \boldsymbol{\omega}_k, \mathbf{0} \right)
% \end{equation}
% where $(\hat{\cdot})$ denotes the estimated value. Concurrently, the error-state covariance matrix $\mathbf{P}$ is propagated as:
% \begin{equation} \label{eq:cov_predict}
% \mathbf{P}_{k|k-1} = \boldsymbol{\Phi}_{k-1} \mathbf{P}_{k-1|k-1} {\boldsymbol{\Phi}_{k-1}}^{\top} + \mathbf{Q}_{k-1}
% \end{equation}
% where the error-state transition matrix $\boldsymbol{\Phi}_{k-1}$ and the discrete noise covariance $\mathbf{Q}_{k-1}$ are identically defined as in \cite{mourikis2007multi}.

\textit{1) State Update Without Registered Map Points:} Following the IMU-based state propagation, this stage triangulates up to 100 highly co-visible features within the sliding window. Their optimal indexing resolutions are dynamically derived from depth-dependent projection intercepts via geometric similarity. Subsequently, ray parameters—encompassing the assigned resolution, ray origin, and direction—are transmitted to the edge device, from which the 3D position of the intersected point is received to perform the state update. To ensure continuous observability, the retrieved points undergo a rigorous multi-view consistency check, with triangulated features serving as a robust fallback when valid map associations are insufficient. As this step neither involves registered map points nor optimizes feature positions, the corresponding state vector is defined as follows:
\begin{gather}
\boldsymbol{x}_k^{sub} =
\left[
{\boldsymbol{x}_I}^{\top}\;
{\boldsymbol{x}_C}^{\top}\;
{\boldsymbol{x}_{Y_0}}^{\top}\;
{\boldsymbol{x}_{Y_1}}^{\top}\;
t_I^C
\right]^{\top}
\end{gather}
Consider the nonlinear measurement model
\begin{equation}
\mathbf{z}_{w,k} = h(\boldsymbol{x}_k^{sub}) + \mathbf{n}_{w,k},
\end{equation}
\noindent where $h(\cdot)$ denotes the projection from ray-indexed 3D points to the image plane, $\mathbf{z}_{w,k}$ represents the corresponding 2D observations across frames, and $\mathbf{n}_{w,k} \sim \mathcal{N}(0, \mathbf{R}_{w,k})$ is zero-mean Gaussian noise with diagonal covariance $\mathbf{R}_{w,k}$.

After linearization at $\hat{\mathbf{x}}_{k|k-1}^{sub}$, the state update is given by
\begin{gather}
\hat{\boldsymbol{x}}_{k|k}^{sub} = \hat{\boldsymbol{x}}_{k|k-1}^{sub} \boxplus \mathbf{K}_k ( \mathbf{z}_{w,k} - h(\hat{\boldsymbol{x}}_{k|k-1}^{sub}) )\\
\mathbf{P}_{k|k}^{\prime} = \mathbf{P}_{k|k-1} - \mathbf{K}_k \mathbf{H}_k \mathbf{P}_{k|k-1}\\
\mathbf{K}_k = \mathbf{P}_{k|k-1} \mathbf{H}_k^{\top} \left( \mathbf{H}_k \mathbf{P}_{k|k-1} \mathbf{H}_k^{\top} + \mathbf{R}_{w,k} \right)^{-1}
\end{gather}
\noindent where $\mathbf{H}_k$ is the Jacobian of $h(\cdot)$ with respect to $\mathbf{x}_k^{sub}$.

\textit{2) State Update With Registered Map Points:} To fully exploit the structure information of multi-resolution voxelized map and bound the estimation drift, we perform a subsequent update incorporating the previously registered map points. Once suitable map points are selected, they are utilized to construct robust reprojection constraints with their corresponding 2D observations across the sliding window. The nonlinear measurement model for the full state vector $\boldsymbol{x}_k$ is formulated as:
\begin{equation} \label{eq:map_meas_model}
\mathbf{z}_{m,k} = h(\boldsymbol{x}_k) + \mathbf{n}_{m,k},
\end{equation}
where $m$ is the number of suitable map points, $\mathbf{z}_{m,k}$ represents the 2D feature correspondences of these suitable registered map points, $\mathbf{n}_{m,k}$ denotes the measurement noise $\mathbf{n}_{m,k} \sim \mathcal{N}(\mathbf{0}, \mathbf{R}_{m,k})$, and $\mathbf{R}_{m,k}$ is a diagonal matrix with a certain eigenvalue in our system. Based on the intermediate state $\hat{\boldsymbol{x}}_{k|k}^{\prime}$ and covariance $\mathbf{P}_{k|k}^{\prime}$ derived from the previous feature update, the full state and its covariance are ultimately refined as follows:
\begin{gather}
\hat{\boldsymbol{x}}_{k} = \hat{\boldsymbol{x}}_{k} \boxplus \mathbf{K}_k \left( \mathbf{z}_{m,k} - h(\hat{\boldsymbol{x}}_{k}) \right)\label{eq:map_state_update} \\
\mathbf{P}_{k|k} = \mathbf{P}_{k|k}^{\prime} - \mathbf{K}_k \mathbf{H}_k \mathbf{P}_{k|k}^{\prime}\label{eq:map_cov_update} \\
\mathbf{K}_k = \mathbf{P}_{k|k}^{\prime} \mathbf{H}_k^{\top} \left( \mathbf{H}_k \mathbf{P}_{k|k}^{\prime} \mathbf{H}_k^{\top} + \mathbf{R}_{m,k} \right)^{-1}\label{eq:map_kalman}
\end{gather}
\noindent where $\mathbf{H}_k$ is the measurement Jacobian with respect to the full state $\boldsymbol{x}_k$.

To efficiently establish robust constraints for upcoming frames, upon completion of the state update, the system utilizes the newly refined pose to query the cloud for the 3D positions of new features in the current frame via ray-based indexing. Subsequently, these valid intersected points are dynamically integrated into the registered prior map, serving as a reliable candidate pool for establishing robust map point constraints in subsequent frames.

\textit{3) Suitable Registered Map Points Selection:} Following the registration of newly intersected 3D points, the registered prior map points are maintained via a voxel grid structure \cite{yuan2024sr} to ensure highly efficient point management. The spatial environment is partitioned into voxels of 0.3 × 0.3 × 0.3 m for indoor scenarios and 1.0 × 1.0 × 1.0 m for outdoor scenarios, with a strict capacity limit of five points per voxel. To facilitate multi-frame association and direct memory access, each registered point maintains four fundamental attributes: a unique feature ID, its global 3D position, the host keyframe denoting its origin, and the host voxel. Since the voxels housing these newly registered points inherently reside within the active field of view, map points contained within them and their surroundings are highly likely to provide valid spatial constraints for new frames. Consequently, to strictly bound the computational overhead while ensuring a broad spatial distribution, the selection is expanded to the 2-neighborhood of these active voxels. Ultimately, all map points within this local vicinity are extracted to formulate robust constraints for the subsequent state update.
\begin{table}[t]
\caption{Datasets of All Sequences For Evaluation}
\label{tab:datasets}
\centering
\setlength{\tabcolsep}{12pt}
\begin{tabular}{c c c c}
\toprule % 最顶部的粗线
 & \textbf{Name} & \begin{tabular}[c]{@{}c@{}}\textbf{Duration}\\\textbf{(min:sec)}\end{tabular} & \textbf{Distance} \\
\midrule % 表头下方的细线
\textit{euroc\_1} & V1\_01\_easy & 2:24 & 58.6 m \\
\textit{euroc\_2} & V1\_02\_medium & 1:24 & 75.9 m \\
\textit{euroc\_3} & V1\_03\_difficult & 1:45 & 79.0 m \\
\textit{euroc\_4} & V2\_01\_easy & 1:52 & 36.5 m \\
\textit{euroc\_5} & V2\_02\_medium & 1:55 & 83.2 m \\
\textit{euroc\_6} & V2\_03\_difficult & 1:55 & 86.1 m \\
\midrule % EuRoC 和 TUM-VI 之间的细线
\textit{kaist\_1} & Urban28 & 32:54 & 11.47 km \\
\textit{kaist\_2} & Urban29 & 7:24 & 3.6 km \\
\textit{kaist\_3} & Urban32 & 18:17 & 7.1 km \\
\textit{kaist\_4} & Urban38 & 36:02 & 11.42 km \\
\textit{kaist\_5} & Urban39 & 31:07 & 11.06 km \\
\bottomrule % 最底部的粗线
\end{tabular}
\vspace{-0.4cm} % 添加负间距,数值可以根据视觉效果微调（如 -0.2cm 到 -0.6cm）
\end{table}

\textit{4) Voxel Resolution Selection:} To guarantee robust cross-modal data association, the retrieved voxel resolution must strictly align with the physical footprint of the 2D observation. Specifically, undersized voxels render the indexing results highly sensitive to pose errors, whereas oversized voxels blur structural details and compromise association precision, as experimentally verified in Sec. V-B. Through geometric similarity, this real-world scale is explicitly determined by the depth-projected intercept of the camera pixel. Consequently, the optimal voxel resolution $r^*$ is dynamically assigned by minimizing the difference between the available map resolutions and the physical pixel footprint:
\begin{equation} \label{eq:voxel_res}
r^* = \arg \min_{r \in \mathcal{R}} \left| r - \frac{d}{f} \right|
\end{equation}
\noindent where $d$ is the estimated depth of the feature point in the camera frame, $f$ represents the camera's focal length, and $\mathcal{R}$ denotes the set of available voxel resolutions.

% \subsection{Point Cloud Indexing}
% To efficiently traverse the voxelized prior map, the 3D-DDA algorithm is employed. The ray search dynamically expands from the triangulated depth along the line of sight, continuing until it either intersects a valid 3D point or exceeds the maximum indexing range. To guarantee robust 2D-to-3D correspondences, retrieved candidates are subjected to a rigorous multi-view consistency check. Furthermore, to maintain continuous state observability, purely triangulated points are leveraged as a robust fallback in scenarios where valid map associations are insufficient.

\subsection{Communication Payload and Latency Analysis}
To substantiate the claim of minimal data transmission between the cloud server and the edge device, we provide a quantitative analysis of the communication payload. Unlike other cloud-assisted VIO methods that transmit heavy local maps or raw images, our system communicates only compact, parameterized geometric primitives.
\subsubsection{Data Volume Analysis}
% The bidirectional data flow is summarized as follows:
\begin{itemize}
    \item \textbf{Uplink Payload (Edge-to-Cloud):} For each feature requiring a 3D position query, the edge device transmits a set of ray parameters, including the ray origin ($\mathbf{x}$), initial distance ($d$), ray direction ($\mathbf{v}$), and the selected resolution level ($r$). Represented by single-precision floating-point numbers, these four parameters result in a compact 32-byte payload per feature.
    \item \textbf{Downlink Payload (Cloud-to-Edge):} Upon a successful hit via the 3D-DDA algorithm, the cloud server returns only the global 3D position ($x, y, z$) of the intersected map point. This constitutes a mere 12-byte payload per retrieved point.
\end{itemize}

Following the configuration in, the system tracks a maximum of 200 keypoints per frame. Under this peak load, the total bidirectional transmission is approximately 0.0084 MB per frame. It is worth noting that the actual data volume in practical operation is typically much smaller because 3D position queries are only necessitated by newly observed or unindexed features. Nevertheless, we utilize this 200-point scenario as a conservative baseline for all subsequent analysis to ensure a robust evaluation of the system performance under extreme conditions.

\subsubsection{Real-time Feasibility}
To evaluate practical feasibility, we consider a typical industrial edge computing scenario (e.g., Skyworth AR terminals) with a stable bandwidth of 1.66 MB/s. Under the aforementioned maximum load of 200 points, the transmission latency is only 5.1 ms. Combined with the time consumption analysis in Sec. V-C, this design ensures that the total per-frame processing time remains strictly below the image interval even in bandwidth-limited environments, thereby satisfying the real-time requirements for odometry pose estimation.

% \subsection{Registered Prior Map Points Management}
% As shown in Fig. \ref{fig:indexing_diagram4},registered map points are managed using a voxel grid similar to \cite{yuan2024sr}. The space is partitioned into voxels of size 0.3 × 0.3 × 0.3 m for indoor scenes and 1.0 × 1.0 × 1.0 m for outdoor scenes, with up to five points stored per voxel. Each point contains four attributes: feature ID (for multi-frame association), global 3D position, host keyframe (where it was first generated), and host voxel (for direct voxel access). This structure enables efficient map point management.After the two-stage state update, the new feature correspondences are triangulated using the updated pose. The appropriate indexing resolution is then determined, and ray-based indexing is performed to retrieve the corresponding 3D points. The ray-hit points are subsequently registered into the voxel map as newly registered map points.
 \begin{table}[h]
% 标题首字母大写,IEEE 模板会自动将其渲染为规范的 Small Caps (小型大写字母)
% 如果你需要和截图中一样的蓝色,可以解除下一行 \color{blue} 的注释
% \color{blue} 
\caption{RMSE of ATE Comparison With State-of-the-Arts (Unit:m)}
\label{tab:ate_comparison}
\centering
% \setlength{\tabcolsep}{5pt}
% c | c c c c c | c 表示在第1列和第2列之间,以及第6列和第7列之间加上竖线
\begin{tabular}{c | c c c c c | c}
\hline
 & \begin{tabular}[c]{@{}c@{}}\textbf{MSCKF}\\ \textbf{W/ Map}\end{tabular} 
 & \begin{tabular}[c]{@{}c@{}}\textbf{GMM}\\ \textbf{W/ Map}\end{tabular} 
 & \begin{tabular}[c]{@{}c@{}}\textbf{Open-}\\\textbf{VINs}\end{tabular} 
 & \begin{tabular}[c]{@{}c@{}}\textbf{VINs-}\\\textbf{Fusion}\end{tabular}
 & \begin{tabular}[c]{@{}c@{}}\textbf{Voxel-}\\\textbf{SVIO}\end{tabular} 
 & \begin{tabular}[c]{@{}c@{}}\textbf{Ours}\end{tabular} \\
\hline
\textit{euroc\_1} & 0.056 & 0.074 & 0.059 & 0.113 & 0.062 & \textbf{0.053} \\
\textit{euroc\_2} & 0.055 & 0.068 & 0.076 & 0.108 & 0.071 & \textbf{0.061} \\
\textit{euroc\_3} & 0.087 & 0.065 & 0.062 & 0.112 & 0.058 & \textbf{0.048} \\
\textit{euroc\_4} & 0.069 & 0.067 & 0.057 & 0.132 & 0.056 & \textbf{0.042} \\
\textit{euroc\_5} & 0.089 & 0.060 & 0.064 & 0.124 & 0.059 & \textbf{0.042} \\
\textit{euroc\_6} & 0.149 & 0.497 & \textbf{0.109} & 0.318 & 0.127 & 0.113 \\
\textit{kaist\_1} & $\times$ & $\times$ & 12.19 & 24.37 & 11.79 & \textbf{10.57} \\
\textit{kaist\_2} & $\times$ & $\times$ & 9.24 & 30.12 & 7.02 & \textbf{6.47} \\
\textit{kaist\_3} & $\times$ & $\times$ & 12.64 & 35.39 & 10.07 & \textbf{9.51} \\
\textit{kaist\_4} & $\times$ & $\times$ & 12.14 & $\times$ & 10.32 & \textbf{9.29} \\
\textit{kaist\_5} & $\times$ & $\times$ & 12.42 & $\times$ & 10.18 & \textbf{8.37} \\
\hline
% 核心底部注释代码：合并7列,使用 p 控制宽度自动换行,并使用 \small 缩小字号
\multicolumn{7}{p{1.00\columnwidth}}{\vspace{2pt} \small \textbf{Denotations:} ``$\times$'' indicates tracking failure or source code unavailability. \textbf{Bold} denotes the 1st-ranked method. ``w/ map'' means using the real-time estimated pose to obtain point clouds from the prior map.} \\
\end{tabular}
\vspace{-0.6cm} % 添加负间距,数值可以根据视觉效果微调（如 -0.2cm 到 -0.6cm）
\end{table}

\section{EXPERIMENTS}
\label{EXPERIMENTS}
% We evaluate MR-Voxel-SVIO on two public benchmark datasets, EuRoC MAV \cite{burri2016euroc} and KAIST \cite{jeong2019complex}, which cover flying drones, and autonomous driving scenarios in both indoor and outdoor environments.There are 11 sequences in the Euroc datasets, but we only use the ViconRoom sequences because only the ViconRoom sequences have the prior LIDAR map. All evaluated sequences (including their names, durations, and traveled distances) are summarized in Table~\ref{tab:datasets}. The Absolute Trajectory Error (ATE) is adopted as the evaluation metric. All experiments are conducted on a workstation equipped with an Intel Core i7-12700KF CPU and 32 GB of RAM.
We evaluate MR-Voxel-SVIO on two public datasets, EuRoC MAV and KAIST, which cover flying drones, and autonomous driving scenarios in both indoor and outdoor environments. There are 11 sequences in the Euroc datasets, but we only use the ViconRoom sequences because only the ViconRoom sequences have the prior LIDAR map. All evaluated sequences (including their names, durations, and traveled distances) are summarized in Table~\ref{tab:datasets}. The Absolute Trajectory Error (ATE) is adopted as the evaluation metric. All experiments are conducted on a workstation equipped with an Intel Core i7-12700KF CPU and 32 GB of RAM.

\subsection{Comparison of the State-of-The-Arts}
We compare our method with several state-of-the-art methods, including MSCKF (w/map) \cite{zuo2019visual}, GMM-Loc \cite{huang2020geometric}, Open-VINS \cite{geneva2020openvins}, VINS-Fusion \cite{qin2018vins}, Voxel-SVIO \cite{yuan2025voxel}. Note that the state-of-the-art ORB-SLAM3 \cite{campos2021orb} is not reported in this section. Since its re-localization and loop closure cannot be disabled, early poses are retroactively optimized by later observations, making a direct comparison unfair. For a fair comparison, results of the open-source systems are obtained using author-provided codes and averaged over five runs, while the results for the closed-source MSCKF (w/map) are directly cited from its paper. Additionally, our method requires global localization poses as inputs. Following \cite{bao2022robust}, we ensure a fair comparison by using ground-truth poses corrupted with translation noise (mean: 2 cm, variance: 0.25 cm) as the default inputs.

As presented in Table~\ref{tab:ate_comparison}, our MR-Voxel-SVIO outperforms the baseline Voxel-SVIO thanks to the integration of the multi-resolution voxelized prior map. Furthermore, it achieves superior accuracy over state-of-the-art methods, yielding a lower Absolute Trajectory Error (ATE) across the majority of the evaluated sequences, and successfully processes all of them, demonstrating its robustness.
\begin{table}[t]
\caption{Comparison of ATE RMSE Using Estimated and Ground-Truth Poses for Prior Map Queries (Unit: m)}
\label{tab:gt_evaluation}
\centering
% 稍微调大一点列间距,让表格在还没填数据时也不会显得太挤
\setlength{\tabcolsep}{13pt} 
\begin{tabular}{c | c c | c c}
\hline
 % 加上了 \textbf 强制加粗
 & \multicolumn{2}{c|}{\textbf{w/ map}} & \multicolumn{2}{c}{\textbf{w/ GT}} \\
 % 建议把子表头也加上加粗,看起来更协调
 & GMM & Ours & GMM & Ours \\
\hline
\textit{euroc\_1} & 0.074 & \textbf{0.053} & 0.071 & \textbf{0.045} \\
\textit{euroc\_2} & 0.068 & \textbf{0.061} & 0.059 & \textbf{0.047} \\
\textit{euroc\_3} & 0.065 & \textbf{0.048} & 0.062 &\textbf{0.041} \\
\textit{euroc\_4} & 0.067 & \textbf{0.042} & 0.060 & \textbf{0.035} \\
\textit{euroc\_5} & 0.060 & \textbf{0.042} & 0.055 & \textbf{0.039} \\
\textit{euroc\_6} & 0.497 & \textbf{0.113} & 0.258 & \textbf{0.095} \\
\textit{kaist\_1} & $\times$ & \textbf{10.57} & $\times$ & \textbf{10.02} \\
\textit{kaist\_2} & $\times$ & \textbf{6.47} & $\times$ & \textbf{5.76} \\
\textit{kaist\_3} & $\times$ & \textbf{9.51} & $\times$ & \textbf{7.58} \\
\textit{kaist\_4} & $\times$ & \textbf{9.29} & $\times$ & \textbf{8.64} \\
\textit{kaist\_5} & $\times$ & \textbf{8.37} & $\times$ & \textbf{6.98} \\
\hline
\end{tabular}

% --- 核心优化：把长注释用 minipage 独立放在表格外面 ---
\vspace{4pt}
\begin{minipage}{0.95\columnwidth}
\small \textbf{Denotations:} ``$\times$'' means the open-source system fails to run entirely on the corresponding sequence. ``w/ map'' means using the real-time estimated pose to obtain point clouds from the prior map. ``w/ GT'' means using the ground-truth pose to obtain point clouds from the prior map. \textbf{Bold} values indicate the best performance.
\end{minipage}
\vspace{-0.4cm} % 添加负间距,数值可以根据视觉效果微调（如 -0.2cm 到 -0.6cm）
\end{table}

% Although a similar ray-casting approach to \cite{huang2020geometric} is used for prior map point retrieval, the prior map representations and their coupling mechanisms differ significantly. As shown in Table~\ref{tab:gt_evaluation}, our approach provides substantially greater accuracy gains than the GMM-based method.On most sequences, when using ground-truth poses to retrieve map points, our method still outperforms GMM operating with ground-truth-based localization.
To further evaluate the core optimization capabilities of the ray-casting-based methods, we query the prior map using ground-truth poses, and compare our system with GMM-Loc, which employs a similar ray-casting retrieval strategy. As shown in Table~\ref{tab:gt_evaluation}, the ``w/ GT'' configuration eliminates query pose drift to reveal the theoretical performance upper bound of both systems given highly accurate 2D-3D associations. The results demonstrate that our system still significantly outperforms GMM even under these ideal query conditions.

\begin{table}[h]
\caption{Ablation Study of Voxel Resolution on EurocMav Datasets (Unit: m)}
\label{tab:ablation_euroc}
\centering
\setlength{\tabcolsep}{10pt}
\begin{tabular}{c | c c c | c}
\hline
 & \multicolumn{3}{c|}{Fixed Resolution} & Ours \\
 & \textbf{0.01 m} & \textbf{0.02 m} & \textbf{0.03 m} & \textbf{(Multi-Res)} \\
\hline
\textit{euroc\_1} & 0.056 & 0.056 & 0.059 & \textbf{0.053} \\
\textit{euroc\_2} & 0.067 & 0.064 & 0.070 & \textbf{0.061} \\
\textit{euroc\_3} & 0.051 & 0.053 & 0.054 & \textbf{0.048} \\
\textit{euroc\_4} & 0.049 & 0.047 & 0.046 & \textbf{0.042} \\
\textit{euroc\_5} & 0.047 & 0.050 & 0.056 & \textbf{0.042} \\
\textit{euroc\_6} & 0.120 & 0.118 & 0.123 & \textbf{0.113} \\
\hline
\end{tabular}
\vspace{-0.2cm} % 添加负间距,数值可以根据视觉效果微调（如 -0.2cm 到 -0.6cm）
\end{table}

\begin{table}[h]
\caption{Ablation Study of Voxel Resolution on KAIST Datasets (Unit: m)}
\label{tab:ablation_kaist}
\centering
% 注意这里加了一列：c | c c c c | c
\begin{tabular}{c | c c c c | c}
\hline
 & \multicolumn{4}{c|}{Fixed Resolution} & Ours \\
 & \textbf{0.05 m} & \textbf{0.10 m} & \textbf{0.15 m} & \textbf{0.20 m} & \textbf{(Multi-Res)} \\
\hline
\textit{kaist\_1} & 11.06 & 10.72 & 10.89 & 11.02 & \textbf{10.57} \\
\textit{kaist\_2} & 6.77 & 6.59 & 6.98 & 6.67 & \textbf{6.47} \\
\textit{kaist\_3} & 9.98 & 9.74 & 9.69 & 10.02 & \textbf{9.51} \\
\textit{kaist\_4} & 10.11 & 9.46 & 9.82 & 10.04 & \textbf{9.29} \\
\textit{kaist\_5} & 9.54 & 9.32 & 8.68 & 9.04 & \textbf{8.37} \\
\hline
\end{tabular}
\vspace{-0.4cm} % 添加负间距,数值可以根据视觉效果微调（如 -0.2cm 到 -0.6cm）
\end{table}

\subsection{Ablation Study}
\label{Ablation Study}
As discussed in Sec. IV-E.4, adaptively aligning the retrieved voxel resolution with the physical pixel footprint is crucial for robust 2D-3D association. Excessively small voxels make the system highly vulnerable to pose drift, whereas overly large voxels blur structural details and diminish the association precision of the maximum intensity gradients. To investigate this, we evaluate the Absolute Trajectory Error (ATE) using single-resolution maps (0.01, 0.02, and 0.03m for EuRoC MAV; 0.05, 0.10, 0.15, and 0.20m for KAIST) against our multi-resolution strategy. As demonstrated in Table~\ref{tab:ablation_euroc} and Table~\ref{tab:ablation_kaist}, the proposed multi-resolution voxelized prior map effectively enhance the overall pose accuracy.
\begin{figure}[h]
      \centering
      \includegraphics[width=1.0\linewidth, trim=0.3cm 2.6cm 0.5cm 3cm, clip]{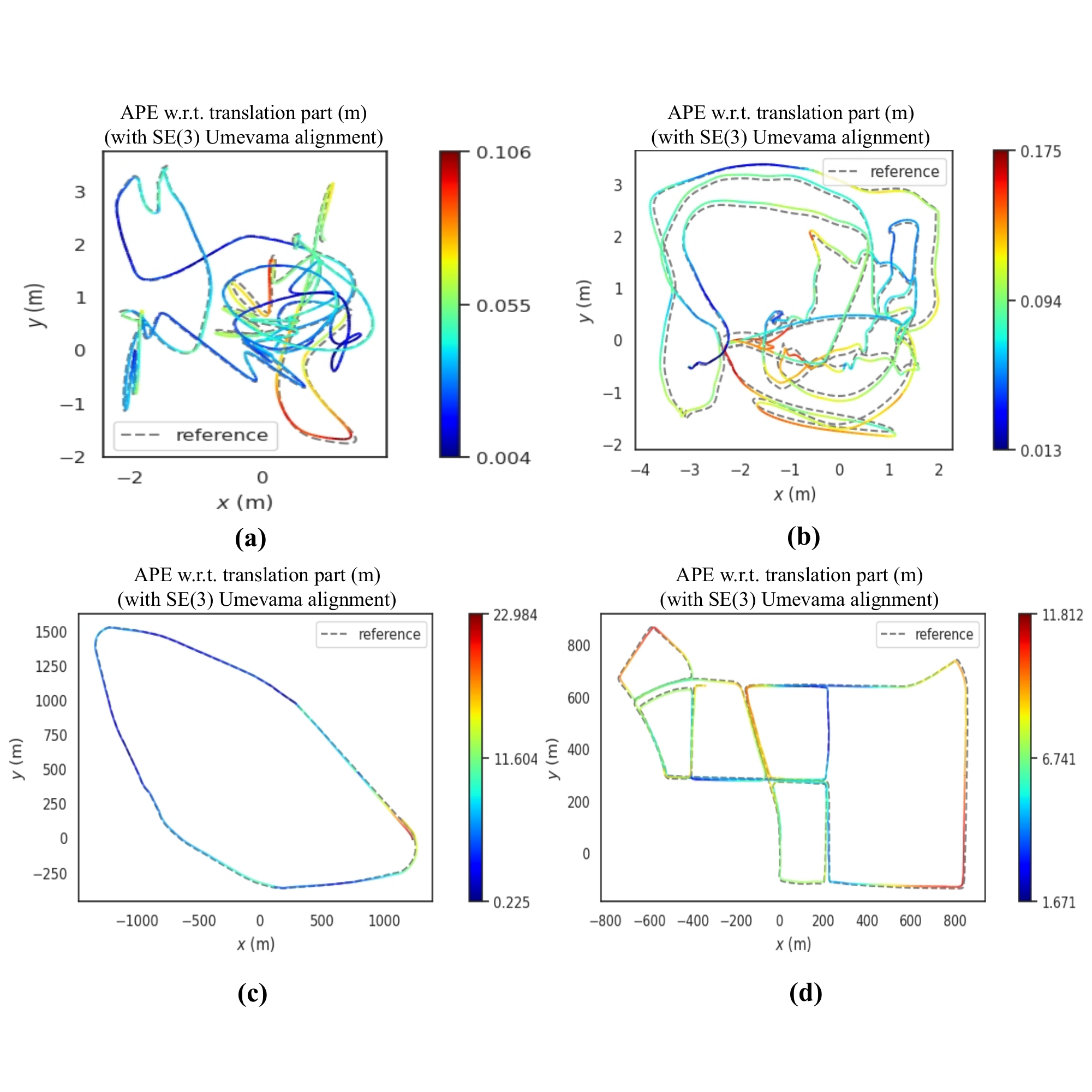}
      \caption{(a)–(d) are the comparison results between our estimated trajectories and ground truth on the exemplar sequences \textit{euroc\_3}, \textit{euroc\_6}, \textit{kaist\_3} and \textit{kaist\_5}.}
      \label{figurelabel}
\vspace{-0.4cm} % 添加负间距,数值可以根据视觉效果微调（如 -0.2cm 到 -0.6cm）
   \end{figure}

\subsection{Time Consumption}
\label{Time Consumption}
Table~\ref{tab:time_consumption} reports the runtime performance of MR-Voxel-SVIO by measuring the average processing time (in milliseconds) of its main components: (a) image processing (converting the stereo image into basic data type and compute the grayscale gradient for each pixel), (b) feature extraction and tracking, (c) state update, (d) suitable registered map points selection, (e) ray indexing, as well as the total per-frame processing time for stereo images. All timings are averaged over entire sequences.

The results demonstrate robust real-time performance across all evaluated sequences. Crucially, as analyzed in Section IV-G, our system incurs a minimal transmission payload, resulting in a latency of only 5.1 ms. Even when accounting for this communication delay, the total per-frame processing time including both local computation and remote map querying remains strictly within the MAT. This validates the practical feasibility of our edge-cloud collaborative architecture, ensuring real-time pose estimation even in bandwidth-limited environments.
\begin{table}[h]
\caption{Time Consumption Per Stereo Image (Unit: ms)}
\label{tab:time_consumption}
\centering
% 可以微调 tabcolsep 让表格整体胖瘦合适,这里设为 6pt (默认值)
\setlength{\tabcolsep}{6pt} 
\begin{tabular}{c | c c c c c c | c}
\hline
 & \textbf{(a)} & \textbf{(b)} & \textbf{(c)} & \textbf{(d)} & \textbf{(e)} & \textbf{Total} & \textbf{MAT} \\
\hline
\textit{euroc\_1}  & 6.74 & 10.76 & 5.21 & 3.31 & 4.86 & 31.76 & 50 \\
\textit{euroc\_2}  & 6.88 & 11.21 & 2.97 & 3.75 & 6.33 & 31.76 & 50 \\
\textit{euroc\_3}  & 7.00 & 11.97 & 2.35 & 3.46 & 5.62 & 30.85 & 50 \\
\textit{euroc\_4}  & 6.95 & 11.09 & 4.73 & 3.28 & 4.08 & 30.94 & 50 \\
\textit{euroc\_5}  & 7.05 & 12.04 & 2.71 & 3.43 & 4.93 & 31.00 & 50 \\
\textit{euroc\_6}  & 8.50 & 14.88 & 1.51 & 2.88 & 5.59 & 33.87 & 50 \\
\textit{kaist\_1}  & 13.51 & 18.89 & 3.05 & 3.22 & 2.49 & 41.76 & 100 \\
\textit{kaist\_2}  & 15.01 & 18.79 & 2.78 & 2.92 & 2.70 & 42.50 & 100 \\
\textit{kaist\_3}  & 13.95 & 18.54 & 2.51 & 3.67 & 2.32 & 42.04 & 100 \\
\textit{kaist\_4}  & 12.52 & 17.95 & 2.54 & 3.35 & 2.45 & 39.16 & 100 \\
\textit{kaist\_5}  & 13.72 & 18.42 & 2.80 & 3.87 & 3.05 & 42.07 & 100 \\
\hline
\end{tabular}

% --- 核心修改：使用 minipage 将注释独立放在表格下方 ---
\vspace{4pt} % 和表格拉开一点点间距
\begin{minipage}{0.98\columnwidth}
\small \textbf{Denotations:} (a) image processing (b) feature extraction and tracking (c) state update (d) suitable registered map points selection and (e) 3D-DDA-based ray traversal. MAT: Maximum available time per frame.The image frequencies of \textit{euroc\_mav},\textit{kaist} are 20\,Hz and 10\,Hz respectively.
\end{minipage}
\vspace{-0.4cm} % 添加负间距,数值可以根据视觉效果微调（如 -0.2cm 到 -0.6cm）
\end{table}

\subsection{Visualization for Trajectory}
As shown in Fig. \ref{figurelabel}, we visualize the comparison between the estimated trajectory and the ground truth, where the estimated trajectory closely aligns with the ground truth.

% \begin{figure}[h]
%       \centering
%       \includegraphics[width=1.0\linewidth, trim=0.3cm 2.6cm 0.5cm 3cm, clip]{figure/fig4.pdf}
%       %\includegraphics[scale=1.0]{figurefile}
%       \caption{(a)–(d) are the comparison results between our estimated trajectories and ground truth on the exemplar sequences \textit{euroc\_3}, \textit{euroc\_6}, \textit{kaist\_3} and \textit{kaist\_5}.}
%       \label{figurelabel}
% \vspace{-0.6cm} % 添加负间距,数值可以根据视觉效果微调（如 -0.2cm 到 -0.6cm）
%    \end{figure}

\section{CONCLUSIONS}
\label{CONCLUSIONS}
This letter proposes a multi-resolution prior map construction method and a corresponding map-based VIO system, MR-Voxel-SVIO, to enable efficient edge-cloud collaboration. By partitioning the prior LiDAR map into multiple resolutions and employing a 3D-DDA-based ray traversal strategy, our system enables precise 2D-to-3D data association while minimizing the volume of data required for transmission. Experimental results on the EuRoC MAV and KAIST datasets demonstrate that the proposed approach achieves state-of-the-art accuracy and robustness across diverse indoor and outdoor scenarios. Future work will focus on integrating a global re-localization module to bound accumulated drift, enabling more accurate 2D-3D associations to further enhance real-time pose estimation accuracy.

% \addtolength{\textheight}{-12cm}   % This command serves to balance the column lengths
%                                   % on the last page of the document manually. It shortens
%                                   % the textheight of the last page by a suitable amount.
%                                   % This command does not take effect until the next page
%                                   % so it should come on the page before the last. Make
%                                   % sure that you do not shorten the textheight too much.

% %%%%%%%%%%%%%%%%%%%%%%%%%%%%%%%%%%%%%%%%%%%%%%%%%%%%%%%%%%%%%%%%%%%%%%%%%%%%%%%%

% %%%%%%%%%%%%%%%%%%%%%%%%%%%%%%%%%%%%%%%%%%%%%%%%%%%%%%%%%%%%%%%%%%%%%%%%%%%%%%%%

% %%%%%%%%%%%%%%%%%%%%%%%%%%%%%%%%%%%%%%%%%%%%%%%%%%%%%%%%%%%%%%%%%%%%%%%%%%%%%%%%

% \begin{thebibliography}{99}

% \bibitem{c1} G. O. Young, ÒSynthetic structure of industrial plastics (Book style with paper title and editor),Ó 	in Plastics, 2nd ed. vol. 3, J. Peters, Ed.  New York: McGraw-Hill, 1964, pp. 15Ð64.

% \end{thebibliography}

\bibliographystyle{IEEEtran} 
\bibliography{IEEEabrv,ref} % 先读官方词典，再读你的文献库

@article{yuan2025voxel,
  title={Voxel-svio: Stereo visual-inertial odometry based on voxel map},
  author={Yuan, Zikang and Lang, Fengtian and Deng, Jie and Luo, Hongcheng and Yang, Xin},
  journal={IEEE Robotics and Automation Letters},
  year={2025},
  publisher={IEEE}
}

@inproceedings{qin2018relocalization,
  title={Relocalization, global optimization and map merging for monocular visual-inertial SLAM},
  author={Qin, Tong and Li, Peiliang and Shen, Shaojie},
  booktitle={2018 IEEE International Conference on Robotics and Automation (ICRA)},
  pages={1197--1204},
  year={2018},
  organization={IEEE}
}

@inproceedings{yamaguchi2020global,
  title={Global-map-registered local visual odometry using on-the-fly pose graph updates},
  author={Yamaguchi, Masahiro and Mori, Shohei and Saito, Hideo and Yachida, Shoji and Shibata, Takashi},
  booktitle={International Conference on Augmented Reality, Virtual Reality and Computer Graphics},
  pages={299--311},
  year={2020},
  organization={Springer}
}

@inproceedings{platinsky2020collaborative,
  title={Collaborative augmented reality on smartphones via life-long city-scale maps},
  author={Platinsky, Lukas and Szabados, Michal and Hlasek, Filip and Hemsley, Ross and Del Pero, Luca and Pancik, Andrej and Baum, Bryan and Grimmett, Hugo and Ondruska, Peter},
  booktitle={2020 IEEE International Symposium on Mixed and Augmented Reality (ISMAR)},
  pages={533--541},
  year={2020},
  organization={IEEE}
}

@article{zuo2019visual,
  title={Visual-inertial localization with prior LiDAR map constraints},
  author={Zuo, Xingxing and Geneva, Patrick and Yang, Yulin and Ye, Wenlong and Liu, Yong and Huang, Guoquan},
  journal={IEEE Robotics and Automation Letters},
  volume={4},
  number={4},
  pages={3394--3401},
  year={2019},
  publisher={IEEE}
}

@inproceedings{caselitz2016monocular,
  title={Monocular camera localization in 3d lidar maps},
  author={Caselitz, Tim and Steder, Bastian and Ruhnke, Michael and Burgard, Wolfram},
  booktitle={2016 IEEE/RSJ International Conference on Intelligent Robots and Systems (IROS)},
  pages={1926--1931},
  year={2016},
  organization={IEEE}
}

@inproceedings{lu2019sharing,
  title={Sharing heterogeneous spatial knowledge: Map fusion between asynchronous monocular vision and lidar or other prior inputs},
  author={Lu, Yan and Lee, Joseph and Yeh, Shu-Hao and Cheng, Hsin-Min and Chen, Baifan and Song, Dezhen},
  booktitle={Robotics Research: The 18th International Symposium ISRR},
  pages={727--741},
  year={2019},
  organization={Springer}
}

@inproceedings{lynen2015get,
  title={Get out of my lab: Large-scale, real-time visual-inertial localization.},
  author={Lynen, Simon and Sattler, Torsten and Bosse, Michael and Hesch, Joel A and Pollefeys, Marc and Siegwart, Roland},
  booktitle={Robotics: Science and Systems},
  volume={1},
  number={10.15607},
  year={2015}
}

@inproceedings{ding2018laser,
  title={Laser map aided visual inertial localization in changing environment},
  author={Ding, Xiaqing and Wang, Yue and Li, Dongxuan and Tang, Li and Yin, Huan and Xiong, Rong},
  booktitle={2018 IEEE/RSJ International Conference on Intelligent Robots and Systems (IROS)},
  pages={4794--4801},
  year={2018},
  organization={IEEE}
}

@inproceedings{kim2018stereo,
  title={Stereo camera localization in 3d lidar maps},
  author={Kim, Youngji and Jeong, Jinyong and Kim, Ayoung},
  booktitle={2018 IEEE/RSJ International Conference on Intelligent Robots and Systems (IROS)},
  pages={1--9},
  year={2018},
  organization={IEEE}
}

@article{huang2020geometric,
  title={Geometric structure aided visual inertial localization},
  author={Huang, Huaiyang and Ye, Haoyang and Jiao, Jianhao and Sun, Yuxiang and Liu, Ming},
  journal={arXiv preprint arXiv:2011.04173},
  year={2020}
}

@article{huang2020gmmloc,
  title={Gmmloc: Structure consistent visual localization with gaussian mixture models},
  author={Huang, Huaiyang and Ye, Haoyang and Sun, Yuxiang and Liu, Ming},
  journal={IEEE Robotics and Automation Letters},
  volume={5},
  number={4},
  pages={5043--5050},
  year={2020},
  publisher={IEEE}
}

@article{bao2022robust,
  title={Robust tightly-coupled visual-inertial odometry with pre-built maps in high latency situations},
  author={Bao, Hujun and Xie, Weijian and Qian, Quanhao and Chen, Danpeng and Zhai, Shangjin and Wang, Nan and Zhang, Guofeng},
  journal={IEEE transactions on visualization and computer graphics},
  volume={28},
  number={5},
  pages={2212--2222},
  year={2022},
  publisher={IEEE}
}

@inproceedings{ye2020monocular,
  title={Monocular direct sparse localization in a prior 3d surfel map},
  author={Ye, Haoyang and Huang, Huaiyang and Liu, Ming},
  booktitle={2020 IEEE International Conference on Robotics and Automation (ICRA)},
  pages={8892--8898},
  year={2020},
  organization={IEEE}
}

@article{belkin2024localization,
  title={Localization of Mobile Robot in Prior 3D {LiDAR} Maps Using Stereo Image Sequence},
  author={Belkin, Ilya V. and Abramenko, Alexander A. and Bezuglyi, Vitaly D. and Yudin, Dmitry A.},
  journal={Computer Optics},
  volume={48},
  number={3},
  pages={406--417},
  year={2024}
}

@inproceedings{geneva2020openvins,
  title={Openvins: A research platform for visual-inertial estimation},
  author={Geneva, Patrick and Eckenhoff, Kevin and Lee, Woosik and Yang, Yulin and Huang, Guoquan},
  booktitle={2020 IEEE International Conference on Robotics and Automation (ICRA)},
  pages={4666--4672},
  year={2020},
  organization={IEEE}
}

@article{qin2018vins,
  title={Vins-mono: A robust and versatile monocular visual-inertial state estimator},
  author={Qin, Tong and Li, Peiliang and Shen, Shaojie},
  journal={IEEE transactions on robotics},
  volume={34},
  number={4},
  pages={1004--1020},
  year={2018},
  publisher={IEEE}
}

@article{campos2021orb,
  title={Orb-slam3: An accurate open-source library for visual, visual--inertial, and multimap slam},
  author={Campos, Carlos and Elvira, Richard and Rodr{\'\i}guez, Juan J G{\'o}mez and Montiel, Jos{\'e} MM and Tard{\'o}s, Juan D},
  journal={IEEE transactions on robotics},
  volume={37},
  number={6},
  pages={1874--1890},
  year={2021},
  publisher={IEEE}
}

@inproceedings{yuan2024sr,
  title={Sr-lio: Lidar-inertial odometry with sweep reconstruction},
  author={Yuan, Zikang and Lang, Fengtian and Xu, Tianle and Yang, Xin},
  booktitle={2024 IEEE/RSJ International Conference on Intelligent Robots and Systems (IROS)},
  pages={7862--7869},
  year={2024},
  organization={IEEE}
}

@article{engel2017direct,
  title={Direct sparse odometry},
  author={Engel, Jakob and Koltun, Vladlen and Cremers, Daniel},
  journal={IEEE transactions on pattern analysis and machine intelligence},
  volume={40},
  number={3},
  pages={611--625},
  year={2017},
  publisher={IEEE}
}

@inproceedings{lucas1981iterative,
  title={An iterative image registration technique with an application to stereo vision},
  author={Lucas, Bruce D and Kanade, Takeo},
  booktitle={IJCAI'81: 7th international joint conference on Artificial intelligence},
  volume={2},
  pages={674--679},
  year={1981}
}

@inproceedings{chen2021rnin,
  title={RNIN-VIO: Robust Neural Inertial Navigation Aided Visual-Inertial Odometry in Challenging Scenes.},
  author={Chen, Danpeng and Wang, Nan and Xu, Runsen and Xie, Weijian and Bao, Hujun and Zhang, Guofeng},
  booktitle={ISMAR},
  pages={275--283},
  year={2021}
}

@inproceedings{amanatides1987fast,
  title={A fast voxel traversal algorithm for ray tracing.},
  author={Amanatides, John and Woo, Andrew and others},
  booktitle={Eurographics},
  volume={87},
  number={3},
  pages={3--10},
  year={1987}
}

@article{jeon2024ecar,
  title={ecar: Edge-assisted collaborative augmented reality framework},
  author={Jeon, Jinwoo and Woo, Woontack},
  journal={arXiv preprint arXiv:2405.06872},
  year={2024}
}

@article{yuan2024sr++,
  title={SR-LIVO: LiDAR-inertial-visual odometry and mapping with sweep reconstruction},
  author={Yuan, Zikang and Deng, Jie and Ming, Ruiye and Lang, Fengtian and Yang, Xin},
  journal={IEEE Robotics and Automation Letters},
  volume={9},
  number={6},
  pages={5110--5117},
  year={2024},
  publisher={IEEE}
}

@article{liu2026gemdepth,
  title={GemDepth: Geometry-Embedded Features for 3D-Consistent Video Depth},
  author={Liu, Yuecheng and Cheng, Junda and Liu, Longliang and Liao, Wenjing and Cheng, Hanrui and Wang, Yuzhou and Yang, Xin},
  journal={arXiv preprint arXiv:2605.10525},
  year={2026}
}

@inproceedings{cheng2025monster,
  title={Monster: Marry monodepth to stereo unleashes power},
  author={Cheng, Junda and Liu, Longliang and Xu, Gangwei and Wang, Xianqi and Zhang, Zhaoxing and Deng, Yong and Zang, Jinliang and Chen, Yurui and Cai, Zhipeng and Yang, Xin},
  booktitle={Proceedings of the Computer Vision and Pattern Recognition Conference},
  pages={6273--6282},
  year={2025}
}

@inproceedings{cheng2024adaptive,
  title={Adaptive fusion of single-view and multi-view depth for autonomous driving},
  author={Cheng, Junda and Yin, Wei and Wang, Kaixuan and Chen, Xiaozhi and Wang, Shijie and Yang, Xin},
  booktitle={Proceedings of the IEEE/CVF Conference on Computer Vision and Pattern Recognition},
  pages={10138--10147},
  year={2024}
}

@article{cheng2024coatrsnet,
  title={Coatrsnet: Fully exploiting convolution and attention for stereo matching by region separation},
  author={Cheng, Junda and Xu, Gangwei and Guo, Peng and Yang, Xin},
  journal={International Journal of Computer Vision},
  volume={132},
  number={1},
  pages={56--73},
  year={2024},
  publisher={Springer}
}

@article{cheng2022region,
  title={Region separable stereo matching},
  author={Cheng, Junda and Yang, Xin and Pu, Yuechuan and Guo, Peng},
  journal={IEEE Transactions on Multimedia},
  volume={25},
  pages={4880--4893},
  year={2022},
  publisher={IEEE}
}

@inproceedings{zhang2025leveraging,
  title={Leveraging consistent spatio-temporal correspondence for robust visual odometry},
  author={Zhang, Zhaoxing and Cheng, Junda and Xu, Gangwei and Wang, Xiaoxiang and Zhang, Can and Yang, Xin},
  booktitle={Proceedings of the AAAI Conference on Artificial Intelligence},
  volume={39},
  number={10},
  pages={10367--10375},
  year={2025}
}

@inproceedings{jia2020d,
  title={D 2 VO: Monocular deep direct visual odometry},
  author={Jia, Qizeng and Pu, Yuechuan and Chen, Jingyu and Cheng, Junda and Liao, Chunyuan and Yang, Xin},
  booktitle={2020 IEEE/RSJ International Conference on Intelligent Robots and Systems (IROS)},
  pages={10158--10165},
  year={2020},
  organization={IEEE}
}

@article{cheng2025monster++,
  title={MonSter++: Unified Stereo Matching, Multi-view Stereo, and Real-time Stereo with Monodepth Priors},
  author={Cheng, Junda and Liao, Wenjing and Cai, Zhipeng and Liu, Longliang and Xu, Gangwei and Wang, Xianqi and Wang, Yuzhou and Yuan, Zikang and Deng, Yong and Zang, Jinliang and others},
  journal={arXiv preprint arXiv:2501.08643},
  year={2025}
}

@article{wei2025decoupling,
  title={Decoupling Bidirectional Geometric Representations of 4D cost volume with 2D convolution},
  author={Wei, Xiaobao and Shu, Changyong and Yue, Zhaokun and Huang, Chang and Liu, Weiwei and Yang, Shuai and Yang, Lirong and Gao, Peng and Zhang, Wenbin and Zhu, Gaochao and others},
  journal={arXiv preprint arXiv:2509.02415},
  year={2025}
}

@article{wei2025wavelet,
  title={A wavelet-based stereo matching framework for solving frequency convergence inconsistency},
  author={Wei, Xiaobao and Liu, Jiawei and Yang, Dongbo and Cheng, Junda and Shu, Changyong and Wang, Wei},
  journal={arXiv preprint arXiv:2505.18024},
  year={2025}
}

\end{document}